# Graph-Based Intercategory and Intermodality Network for Multilabel Classification and Melanoma Diagnosis of Skin Lesions in Dermoscopy and Clinical Images

Xiaohang Fu, Lei Bi, Ashnil Kumar, *Member, IEEE*, Michael Fulham, and Jinman Kim, *Member, IEEE*

*Abstract*—The identification of melanoma involves an integrated analysis of skin lesion images acquired using the clinical and dermoscopy modalities. Dermoscopic images provide a detailed view of the subsurface visual structures that supplement the macroscopic details from clinical images. Melanoma diagnosis is commonly based on the 7-point visual category checklist (7PC), which involves identifying specific characteristics of skin lesions. The 7PC contains intrinsic relationships between categories that can aid classification, such as shared features, correlations, and the contributions of categories towards diagnosis. Manual classification is subjective and prone to intra- and interobserver variability. This presents an opportunity for automated methods to aid in diagnostic decision support. Current state-of-the-art methods focus on a single image modality (either clinical or dermoscopy) and ignore information from the other, or do not fully leverage the complementary information from both modalities. Furthermore, there is not a method to exploit the 'intercategory' relationships in the 7PC. In this study, we address these issues by proposing a graph-based intercategory and intermodality network (GIIN) with two modules. A graph-based relational module (GRM) leverages intercategorical relations, intermodal relations, and prioritises the visual structure details from dermoscopy by encoding category representations in a graph network. The category embedding learning module (CELM) captures representations that are specialised for each category and support the GRM. We show that our modules are effective at enhancing classification performance using three public datasets, and show that our method outperforms state-of-the-art methods at classifying the 7PC categories and diagnosis.

## I. INTRODUCTION

MELANOMA is the deadliest form of skin cancer [1, 2]. It had the fastest rate of incidence increase over the last 50 years compared to all other cancers, with the highest incidence in Australia and New Zealand [1]. Early diagnosis is critical as early stage melanoma is curable [3]. Clinical diagnosis involves assessing and classifying the appearance of a skin lesion. The skin lesion is typically first examined with the naked eye and clinical photography with a digital camera, followed by dermoscopy, and then the two modalities are evaluated together [4, 5]. Clinical images capture a macroscopic view, providing important features including elevation, ulceration, geometry, borders, and colour [6], but they cannot depict subsurface details such as 'dots' and 'globules'. They are also less standardised and there may be large variations in the field-of-view and lighting, and the skin may have tattoos or body piercings. Dermoscopy is more advanced and provides a more comprehensive examination of the lesion, and has been proven to improve the accuracy of melanoma diagnosis [7-10]. Dermoscopy provides a magnified (typically 10×) microscopic view of the lesion by covering the lesion in fluid or using cross-polarised light filters to eliminate surface reflection. This reveals further subsurface structures that cannot be seen on clinical images such as dots, globules, and vascularity, thereby providing valuable visual information. Integrated clinical and dermoscopy examinations that consider both macroscopic and microscopic features are now routine in clinical practice [11]. Examples of clinical and dermoscopic image pairs of skin lesion cases with category classifications are presented in Fig. 1.

| | a) | b) | c) |
|---|---|---|---|
| Dermoscopy Image | | | |
| Clinical Image | | | |
| Diagnosis (DIAG) | Melanoma | Reed or Spitz Nevus | Clark Nevus |
| Pigment Network (PN) | Atypical | Atypical | Absent |
| Streaks (STR) | Irregular | Irregular | Irregular |
| Pigmentation (PIG) | Diffuse Irregular | Absent | Absent |
| Regression Structures (RS) | Combinations | Absent | Absent |
| Dots and Globules (DaG) | Irregular | Irregular | Regular |
| Blue Whitish Veil (BWV) | Present | Present | Absent |
| Vascular Structures (VS) | Absent | Dotted | Dotted |

Fig. 1. Three examples of skin lesions (melanoma, spitz nevus, and clark nevus) showing their appearance on dermoscopic and clinical images, and their category classifications.

The 7-point checklist (7PC) is a reliable and one of the most common diagnostic approaches for melanoma [4, 10, 12]. The method involves simultaneously assessing seven visual characteristics of the lesion: pigment network (PN), streaks (STR), pigmentation (PIG), regression structures (RS), dots and globules (DaG), blue whitish veil (BWV), and vascular structures (VS). Each category contributes to a score if its classification is indicative of melanoma, and the lesion is suspected of being a melanoma if the total score is ≥3 [13]. However, classification of the 7PC is a complex and subjective task as many characteristics are simultaneously assessed, and the expressions of each category can vary (e.g., compare examples of the same class in Fig. 1). This can lead to high intra- and interobserver variations and misdiagnosis [14]. Diagnostic accuracy is also highly dependent on the experience of the examiner [8], and dermoscopic assessment can be difficult to learn [15]. An automated method may improve diagnosis, eliminate subjective evaluation, and be adopted for mass screening.

Early automated methods classified melanoma by first detecting the lesion border, extracting multiple handcrafted features of the lesion relating to shape, texture, and colour, then applied classifiers such as support vector machines (SVMs) [16-21]. Celebi et al. [17] used feature selection algorithms to determine an optimal set of features. Fabbrocini et al. [18] constructed specialised image processing pipelines for each category of the 7PC. However, these handcrafted feature-based methods are heavily dependent on image pre-processing and require specialist knowledge to compute visual descriptors for each category.

More recently, convolutional neural networks (CNNs) have been successful across numerous image-related problems and have been applied to classify skin lesions. Previous efforts have been typically limited to a single modality (most commonly dermoscopy) and directly predicted melanoma diagnosis [22-26]. Progress in this field has been stimulated by the recurring International Skin Imaging Collaboration's skin lesion classification challenge [27, 28]. The high-performing methods in the challenge showed superior accuracy compared to dermatologists for the classification of melanoma. Yu et al. [23] segmented the lesions using a fully convolutional residual network, then performed melanoma classification on cropped lesions. Gessert et al. [26] combined meta-data with dermoscopic images and ensembled multi-resolution CNNs. These methods, however, were designed for single image modality inputs. Their application to multimodal inputs is not trivial, especially in terms of how features from different modalities should be optimally combined.

Multimodal methods for dual clinical and dermoscopy modalities are rarer. Such an approach must address the optimal use of complementary features in the input image pair, account for the value of each image type for different categories, and assimilate the information to improve classification. A common strategy has been the fusion of features from each modality in the late stages of the model [13, 29, 30]. Kawahara et al. [13] performed multiclass skin lesion classification by concatenating different combinations of features extracted from dermoscopic and clinical images by a pretrained Inception network with the patient's meta-data (such as patient gender and lesion location). Yap et al. [29] proposed EmbeddingNet, which concatenated features extracted from clinical and dermoscopic images with meta information for classification. Ge et al. [30] proposed TripleNet, which concatenated features extracted from each modality and fed it into a third subnetwork for classification. The hyper-connected CNN (HcCNN) [31] added a hyper-branch to combine and process information from the dermoscopy and clinical branches at different scales.

However, none of the previous methods explicitly used the relationships between the 7PC categories and diagnosis to derive or leverage complementary features, such as the direct contributions of the categories toward diagnosis. While the 7PC is not directly based on clinical features, there are potential relationships between features in the clinical and dermoscopy modalities that can be leveraged to improve classification accuracy. For example, lesions that appear benign clinically and suspicious in dermoscopy (or vice versa, i.e., lacking clinical-dermoscopic agreement) are treated with extra caution and prompt for a biopsy [32]. Our hypothesis is that the correlations between the presence or absence of different categories, and the exchange of mutual features between the categories, will be valuable in improving classification performance. We propose an architecture to merge these relationships for the 7PC and diagnosis.

Recently, graph neural networks (GNNs) have garnered considerable interest due to their ability to represent relationships between attributes and have proven powerful in a range of applications [33]. GNNs model objects as nodes and capture the relationships between them as edges, and therefore can represent a large variety of data types including non-Euclidean data. Graph convolutional networks (GCNs) [34] extend GNNs by performing convolutions on neighbouring node features. Rather than assume that each neighbour should contribute equally towards the central node, Veličković et al. [35] introduced the graph attention network (GAT) to adaptively assign coefficients for each neighbour. GCNs have been employed to model relations between labels in multilabel image classification problems. Chen et al. [36] modelled interdependencies among the labels as a GCN using word embeddings and combined it with a CNN backbone that extracted visual features. Other frameworks with similar architectures have been explored for multilabel image-based problems [37-41]. Hou et al. [40] combined CNN features of chest X-ray images with disease label embeddings using a transformer encoder, then employed a GCN for visual-semantic learning and multilabel disease classification. Wu et al. [41] used a GCN to characterise the co-occurrence of skin conditions for their classification in clinical images. These existing GCN-based methods, however, are not suitable for multimodal images, as they do not consider how the visual characteristics unique to each modality should be optimally assimilated in the context of deriving intercategorical relationships, or how much influence should the presence or absence of certain features in each modality have on the prediction.

### A. Our Contributions

In this study, we propose a graph-based intercategory and intermodality network (GIIN) for melanoma and 7PC classification that leverages the inherent relationships between visual characteristics that correspond to the 7PC, and complementary information from clinical and dermoscopy images. We leverage this in a manner that prioritises the information from dermoscopy and uses contextual macroscopic

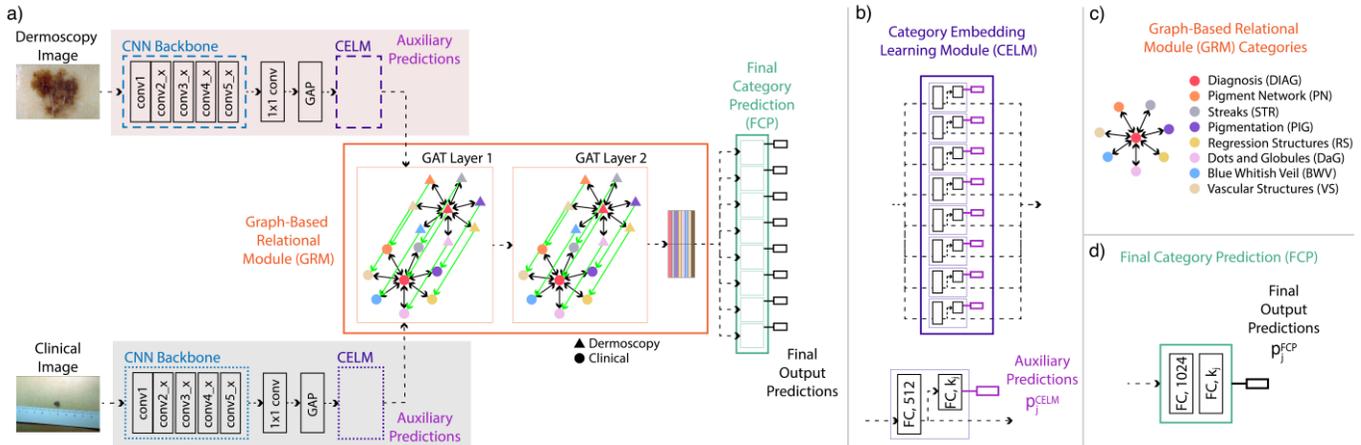

Fig. 2. Overview of our proposed approach: a) the GIIN with CELM and GRM, b) CELM with a detailed isolated unit for a single category, c) star structure of GRM for one modality in which the 7PC categories are connected to the central node of diagnosis, and d) a unit of a final category prediction component. In the Supplementary Materials, we provide a version of this figure with annotated tensor dimensions in Fig. S1, and the adjacency matrix of the GRM in Fig. S2.

information from the clinical images. Our contributions to the state-of-the-art are as follows:

- We propose a graph-based relational module (GRM) to exploit intercategorical relationships. The GRM represents the categories as a set of two star graphs, each corresponding to the dermoscopic or clinical image modality, and is composed of a central diagnosis node with 7 leaves. This topology encodes the relationships between diagnosis and the other 7 categories, and the relationships between the 7 categories. This is different from existing multimodal approaches, as they do not explicitly consider the dependencies between individual labels.

- The GRM includes directed edges between corresponding nodes of the same category in the dermoscopic and clinical star graphs to use complementary visual features from both modalities. These directed edges (from dermoscopy to clinical) prioritise the dermoscopy features and allow the intermodal relationships to self-adapt independently for each category.

- We also propose a category embedding learning module (CELM) to automatically transform general CNN features of the skin lesion from the image pair into embedded representations that are meaningful and specialised to each of the categories. These transformed features are designed to support the relational learning downstream.

We outline these contributions in our overview of the methods in Fig. 2.

## II. METHODOLOGY

### A. Materials

We used the publicly available 7PC dataset [13], which comprises dermoscopic and clinical image pairs from 1,011 patients. The authors split the dataset into 413 training, 203 validation, and 395 test studies. We used the same image splits as the original paper for all our experiments. The size ($h \times w$) of the dermoscopic images ranged from $474 \times 512$ to $532 \times 768$ pixels, and the clinical images ranged from $480 \times 512$ to $532 \times 768$ pixels. There were 8 categories—7 from the 7PC and diagnosis (DIAG). DIAG had 5 different classes: basal cell carcinoma (BCC), nevus (NEV), melanoma (MEL), miscellaneous (MISC), and seborrheic keratosis (SK). The classes of the other categories included: i) absent (ABS), ii) typical (TYP), iii) regular (REG), iv) present (PRS), v) atypical (ATP), and vi) irregular (IR). Classes (iv) to (vi) contribute to the 7-point score. For a full breakdown of the categories including the number of examples of each class, please refer to Table S1 in the Supplementary Materials.

For the ISIC 2017 and 2018 Challenges [27, 42], the attributes included globules (2018 only), milia-like cysts, negative network, pigment network, and streaks. There were 2,000 training and 600 test examples in the 2017 dataset, and 2,594 training and 100 validation examples in the 2018 dataset with publicly available ground truth labels.

### B. Graph-Based Intercategory and Intermodality Network (GIIN)

In our method, information from each modality is first extracted and processed in unique branches by separate CNN backbones and CELMs, then amalgamated in the GRM. The outputs of the GRM are used to predict the 7PC. We use the ResNet-50 model as the CNN backbone for visual feature extraction due to its versatility across a range of image-based problems including skin lesion analysis [22, 24, 29, 31]. Our method can support other backbones and we later show that the CELM and GRM can enhance the performance of various backbone CNNs. We inserted a $1 \times 1$ convolution layer immediately after the 'conv5_x' block of the backbone to reduce the number of feature channels from 2,048 to 512 for each branch. This further reduces the number of parameters downstream in the model and minimises the risk of overfitting. The feature volumes are subsequently vectorised by global average pooling (GAP) layers, producing vectors of $\mathbf{x^{g,D}} \in \mathbb{R}^{1 \times 1 \times 512}$ for dermoscopic images and $\mathbf{x^{g,C}} \in \mathbb{R}^{1 \times 1 \times 512}$ for clinical images.

### C. Category Embedding Learning Module (CELM)

Prior to relational learning, the extracted visual information is transformed into latent feature vectors that are specialised for each category by the CELM. The CELM outputs auxiliary predictions that are used to compute auxiliary cross-entropy losses $L^{CELM,D}$ and $L^{CELM,C}$ for each category and modality that contribute towards the total training loss (details in Section 2.5). There are eight units in the CELM of each modality, corresponding to the 7PC categories and diagnosis. We define the units to output auxiliary classification predictions as:

$$\mathbf{h}_j^{\mathbf{CELM,D}}(\mathbf{x}^{g,D}) = \mathbf{W}_{0,j}^{\mathbf{D}} \mathbf{x}^{g,D} + \mathbf{b}_{0,j}^{\mathbf{D}}$$
$$\mathbf{h}_j^{\mathbf{CELM,C}}(\mathbf{x}^{g,C}) = \mathbf{W}_{0,j}^{\mathbf{C}} \mathbf{x}^{g,C} + \mathbf{b}_{0,j}^{\mathbf{C}} \qquad (1)$$
$$\mathbf{p}_j^{\mathbf{CELM,D}}(\mathbf{h}_j^{\mathbf{CELM,D}}) = \sigma(\mathbf{W}_{1,j}^{\mathbf{D}} \mathbf{h}_j^{\mathbf{CELM,D}} + \mathbf{b}_{1,j}^{\mathbf{D}})$$
$$\mathbf{p}_j^{\mathbf{CELM,C}}(\mathbf{h}_j^{\mathbf{CELM,C}}) = \sigma(\mathbf{W}_{1,j}^{\mathbf{C}} \mathbf{h}_j^{\mathbf{CELM,C}} + \mathbf{b}_{1,j}^{\mathbf{C}}) \qquad (2)$$

where $\mathbf{p}_j^{\mathbf{CELM}} \in \mathbb{R}^{1 \times k}$ denotes the auxiliary predicted class probabilities of category $j$ with $k$ classes, $\sigma$ is softmax activation, and $\mathbf{h}_j^{\mathbf{CELM}} \in \mathbb{R}^{512 \times 1}$ is the intermediate CELM feature vector, with **D** referring to dermoscopy and **C** clinical. The CELM unit of a category is characterised by a set of trainable weights and biases of two fully connected layers, comprising of $\mathbf{W}_{0,j} \in \mathbb{R}^{512 \times 512}$, $\mathbf{W}_{1,j} \in \mathbb{R}^{512 \times k}$, $\mathbf{b}_{0,j} \in \mathbb{R}^{512 \times 1}$, and $\mathbf{b}_{1,j} \in \mathbb{R}^{1 \times k}$.

By carrying out auxiliary predictions and minimising the auxiliary losses, CELM ensures that $\mathbf{h}^{\mathbf{CELM,D}}$ and $\mathbf{h}^{\mathbf{CELM,C}}$ are meaningful for each of the categories prior to relational processing in the GRM. The auxiliary predictions are only used during model training and have no role at inference time.

### D. Graph-Based Relational Module (GRM)

The GRM is composed of two GNN layers with the same graph topologies. Relations between categories are modelled with a star-shaped graph that reflects the inherent associations of the 7PC, consisting of a central diagnosis node that is connected to all other categories via bidirectional (inverse) edges. There are two star-shaped structures in each GNN layer, corresponding to the two modalities. Directed edges propagate information from dermoscopy to the clinical modality. The adjacency matrix is provided in Supplementary Fig. S2.

The input feature vector for the node of category $j$ is $\mathbf{h}_j^{\mathbf{CELM,D}}$ for dermoscopy or $\mathbf{h}_j^{\mathbf{CELM,C}}$ for clinical images. In GCNs, each node receives the aggregated features of its neighbourhood, and a nonlinear activation function is applied on the output. Multiple GCN layers can be stacked to learn higher order representations and allow each node to receive information from more distant nodes (more 'hops') as shown in Fig. 3. In a standard GCN, the neighbouring nodes are weighted equally towards the central node. In the GRM, we assign different contributions for each neighbour.

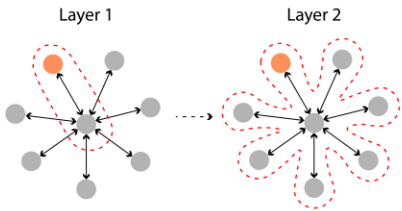

Fig. 3. GCN layers are stacked so that each node receives information from other connected nodes that are a larger number of hops (distance between the nodes) away. In Layer 1, each 7PC node is only connected

to the central diagnosis node. In Layer 2, with an additional hop, the 7PC nodes are also connected. This figure illustrates the star-shaped graph for one modality.

#### 1) GRM Formulation

We used the GAT [35] approach to operate on the graph layers of the GRM, that had topologies and connections as described above. The graph attentional layer computes normalised attention coefficients via a self-attention mechanism:

$$\alpha_{uv} = \frac{\exp(\text{LeakyReLU}(\mathbf{a}^T[\mathbf{W}\mathbf{h}_u \,\|\, \mathbf{W}\mathbf{h}_v]))}{\sum_{k \in N_u} \exp(\text{LeakyReLU}(\mathbf{a}^T[\mathbf{W}\mathbf{h}_u \,\|\, \mathbf{W}\mathbf{h}_k]))} \qquad (3)$$

where $\mathbf{h} = \{\mathbf{h}_1, \mathbf{h}_2, ..., \mathbf{h}_N\}$, $\mathbf{h}_i \in \mathbb{R}^F$ represents the input feature vectors to the GAT layer with $N = 16$, corresponding to the eight categories of the two modalities (i.e., $\mathbf{h}^{\mathbf{CELM,D}} \in \mathbb{R}^{512 \times 8}$ and $\mathbf{h}^{\mathbf{CELM,C}} \in \mathbb{R}^{512 \times 8}$ from Equation 1), $\alpha_{uv}$ is the attention coefficient for node $v$ to $u$, and $N_u$ denotes the neighbourhood nodes of $u$. The feature vectors are linearly transformed via $\mathbf{W} \in \mathbb{R}^{F' \times F}$, the weight matrix shared for every node, then by $\mathbf{a} \in \mathbb{R}^{2F'}$, the weight vector of a single-layer feedforward neural network of the attention mechanism, with $\|$ representing concatenation. LeakyReLU nonlinearity (with a slope of 0.2 for inputs < 0) followed by softmax normalisation are then applied.

The updated feature vectors, $\mathbf{h}' = \{\mathbf{h}'_1, \mathbf{h}'_2, ..., \mathbf{h}'_N\}$, $\mathbf{h}'_i \in \mathbb{R}^{F'}$, can then be computed by proportionally aggregating over neighbouring nodes:

$$\mathbf{h}'_u = \sigma\left(\sum_{v \in N_u} \alpha_{uv} \mathbf{W}\mathbf{h}_v\right) \qquad (4)$$

where $\sigma$ is the exponential linear unit (ELU) nonlinearity [43].

Furthermore, we used the multi-head attention mechanism for better stability, computing multiple independent attention coefficients for each node that are aggregated via concatenation:

$$\mathbf{h}'_u = \|_{m=1}^M \sigma\left(\sum_{v \in N_u} \alpha_{uv}^m \mathbf{W}^m \mathbf{h}_v\right) \qquad (5)$$

where, for the $m$-th attention head, $\alpha_{uv}^m$ are the normalised attention coefficients and $\mathbf{W}^m$ are the weights of the input linear transformation. In the first layer, we use 8 attention heads per node and $F_0 = 8$, thereby computing a total of 64 features, as done in the original paper [35]. For the second layer, we compute $F_1 = 512$ features with a single attention head.

### E. Final Category Classification and Training

The output node vectors of the second GRM layer represent amalgamated complementary features from the two modalities. The pairs of output node vectors that correspond to the same category or diagnosis are then concatenated together. This concatenation yields eight vectors (each with dimensionality $1,024 \times 1$), as there is one vector each from the dermoscopy and dermoscopy-clinical nodes of the GRM. These vectors are fed through the final category prediction (FCP) units. There are eight such units, corresponding to the 7PC categories and diagnosis, each of which comprises two fully connected layers and softmax activation to output class probabilities. The model is trained by minimising the sum of the cross-entropy losses between the final predictions and ground truth classes, and the auxiliary losses from the CELM of both modalities:

$$\arg\min_{\theta} \sum_{i=1}^{M} \sum_{j=1}^{N} L^{total}(d_i, c_i; p_{i,j}^{CELM,D}, p_{i,j}^{CELM,C}, p_{i,j}^{FCP}, y_{i,j}) \quad (6)$$

$$L^{total} = \lambda_D L^{CELM,D} + \lambda_C L^{CELM,C} + L^{FCP} \quad (7)$$

where $y_{i,j}$ denotes the ground truth for category $j$ of case $i$, $p^{FCP}$ is the final category predictions, and $p^{CELM,D}$ and $p^{CELM,C}$ are the auxiliary predictions. $L^{total}$ is the total cross-entropy loss, $d$ the dermoscopic image, $c$ the clinical image, and $\theta$ represents the model parameters. $L^{total}$ comprises $L^{FCP}$ and the CELM auxiliary losses $L^{CELM,D}$ and $L^{CELM,C}$, which are balanced against $L^{FCP}$ with $\lambda_D$ and $\lambda_C$.

### F. Implementation Details

The following implementation and hyperparameter choices were kept consistent for all experiments to ensure that there was a fair comparison. The networks were trained end-to-end from scratch for 100 epochs with a batch size of 4. We employed the Adam optimiser [44] to minimise the mean multi-task cross-entropy loss at a fixed learning rate of 0.00001, with a first moment estimate of 0.9 and second moment estimate of 0.999. We set the weights of the auxiliary losses, $\lambda_D$ and $\lambda_C$, to both be 0.5, as determined empirically (evaluation in Table S2). The weights of the backbone were initialised using the ResNet-50 model that was pretrained on ImageNet [45], then further trained with the rest of our model. This approach has been commonly adopted in previous skin lesion studies to overcome the limitations of using a small training set [13, 22, 23, 26, 31]. Weights of the $1 \times 1$ convolutional layers were initialised using He et al.'s method [46], while biases were initialised to zero. Glorot (Xavier) initialisation [47] was used for fully connected layers. Dropout was not used in any experiment.

Each image was mean-subtracted and normalised to unit variance using the training set mean and standard deviation of its image type (dermoscopy or clinical). We employed standard online (on-the-fly) image data augmentation by randomly applying a flip (horizontal or vertical), rotation (of 90, 180 or 270 degrees), random crop, or adding Gaussian noise (between 0 and 0.2*255) to the input images. All images were resized to $512 \times 768$ pixels with bilinear interpolation, to match the size of the largest images in the dataset. The order of training examples was shuffled for every epoch. All networks were implemented using the PyTorch framework [48]. Both training and testing were performed with a 12GB NVIDIA GTX Titan V GPU. Training required about 4 hours for completion.

### G. Evaluation Setup

Our performance metrics were the commonly used area under receiver operating characteristic curve (AUC), sensitivity (Sens), specificity (Spec), and precision (Prec). For AUC, we followed the one-against-all approach as adopted by comparison studies [13, 31].

#### 1) Comparison to the State-of-the-Art

We benchmarked the proposed GIIN against state-of-the-art methods using the same dataset. The methods included Inception-unbalanced [13], Inception-balanced [13], Inception-combined [13], EmbeddingNet [29], TripleNet [30], and HcCNN [31].

#### 2) Ablation Study

We performed an ablation study to determine the contributions from each module. The baseline was the version of our GIIN without CELM and GRM. We then evaluated the effectiveness of CELM by incorporating it into the branches of each modality. We assessed the performance of several topological variants of GRM (Fig. 4) to verify the optimal directionality of the intermodal edges. The direction of the edges between dermoscopy and clinical features determine which modality has relatively greater influence on classification. We compared this against: i) using separate graphs for each modality without intermodal edges (GRM-Separate), ii) pre-concatenation of the feature vectors for each modality that effectively fused the GRM into a single modality graph (GRM-Fused), iii) using inverse intermodality edges between the modalities (GRM-INV), or iv) edges directed from clinical to dermoscopy (GRM-CD). A simple concatenation of feature vectors from both modalities without GRM was applied

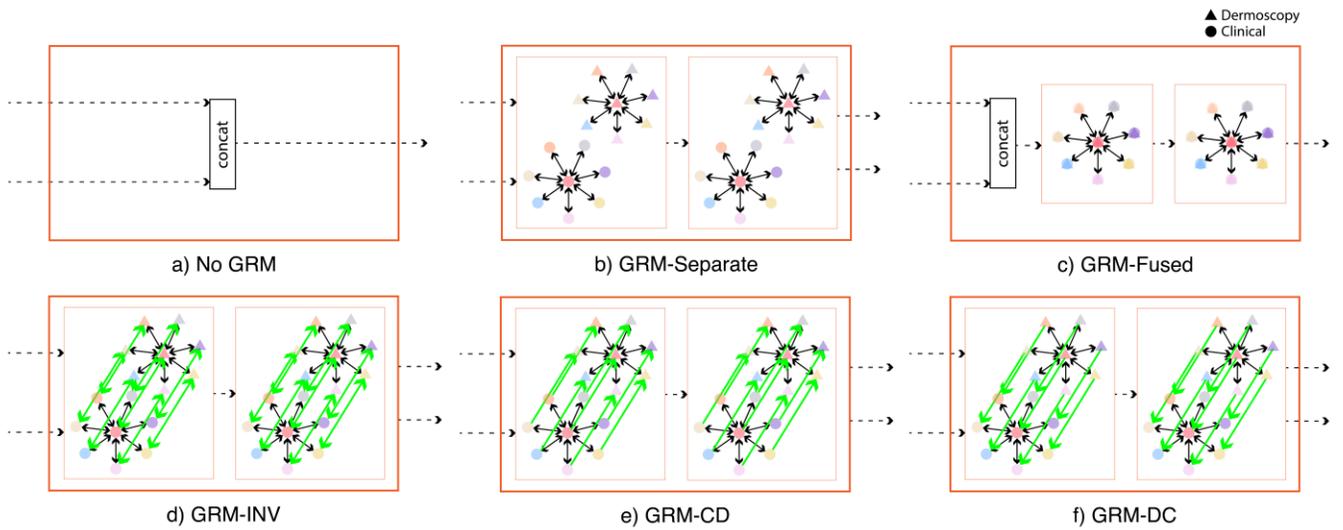

Fig. 4. Different approaches to assimilate multimodal features using a) a simple concatenation without GRM; or variations of our GRM involving b) separate disconnected graphs for each modality; c) pre-concatenation of the feature vectors for each modality; or intermodality edges directed d) inversely between the modalities, e) from clinical to dermoscopy, or f) dermoscopy to clinical (proposed).

for the baseline without CELM and GRM, and for the version that only had CELM.

We evaluated the generalisability of our CELM and GRM using the publicly available ISIC 2017 and ISIC 2018 Challenge datasets [27, 42, 49]. Task 2 concerned superpixel annotation or segmentation of multiple attributes in dermoscopic images. We adopted the tasks for our model by converting them to binary classification, i.e., predicting the presence or absence of each attribute in the image. Our model was modified to suit the single modality inputs of these two datasets by removing the clinical pathway, using one star-shaped graph structure for the one modality, and removing the concatenation step before final predictions. All hyperparameter choices were kept the same. We evaluated the GIIN without either module, with the addition of CELM, and with both CELM and GRM.

We also assessed the robustness of our GIIN to variations of each input image type by data augmentation techniques (flips, rotations, crops, and random Gaussian noise). Each type of augmentation was applied to all the images of one modality while the other modality was kept unaltered. The model used was the best performing model from the ablation study on the 7PC dataset; it was not fine-tuned for this task.

### 3) Application to Different Backbones

We also investigated the applicability of our proposed CELM and GRM to other feature extractor backbones, including MobileNetV2 [50], ResNet-18, and ResNet-34 [51], by swapping out the ResNet-50 backbone. The final output classification layers of each backbone were discarded as they served no purpose. All these alternatives were pretrained on ImageNet and fine-tuned with the rest of our model, as done with our primary backbone.

## III. RESULTS

### A. Comparison to the State-of-the-Art

We compared our GIIN with CELM and GRM-DC to the state-of-the-art methods (Table 1) Our method had an overall superior performance across AUC and average precision taken across all 8 categories, and was second-best across specificity.

### B. Ablation Study

The ablation study showed that CELM and GRM-DC enhanced classification performance of the baseline model (Table 2). The different GRM topologies affected performance differently (Table 3). Note that we also experimented using GRM without CELM, but we found that loss did not decrease during training and the model did not learn successfully. When compared to the baseline with CELM, average AUC was slightly diminished by further including GRM-Fused or GRM-CD. The other three configurations—GRM-Separate, GRM-INV, and our proposed GRM-DC—successfully increased performance, and the largest improvements were found with GRM-DC. Of the three kinds of intermodality edges, the performance of GRM-INV was in close approximation to the average of the directed variants.

The CELM and GRM consistently improved the average performance over the baseline models on the ISIC 2017 and ISIC 2018 datasets (Table S4), as CELM provided an average gain of 1.6 AUC, while GRM added an additional 1.55 AUC. We also include single modality results in Table S5.

TABLE I
CLASSIFICATION PERFORMANCE OF OUR GIIN WITH CELM AND GRM-DC COMPARED TO THE STATE-OF-THE-ART

| | Method | DIAG MEL | PN ATP | STR IR | PIG IR | RS PRS | DaG IR | BWV PRS | VS IR | Avg |
|---|---|---|---|---|---|---|---|---|---|---|
| **AUC** | Inception-unbalanced [13] | 83.2 | 78.6 | 78.3 | 78.1 | 79.9 | 76.4 | 87.0 | 73.4 | 79.4 |
| | Inception-balanced [13] | 84.2 | 78.9 | 78.7 | 79.4 | **83.5** | 78.0 | 87.5 | 76.1 | 80.8 |
| | Inception-combined [13] | 86.3 | 79.9 | 78.9 | 79.0 | 82.9 | 79.9 | 89.2 | 76.1 | 81.5 |
| | EmbeddingNet [29] | 82.5 | 74.5 | 77.7 | 77.9 | 71.3 | 78.5 | 84.8 | 76.9 | 78.0 |
| | TripleNet [30] | 81.2 | 73.8 | 76.3 | 77.6 | 76.0 | 76.0 | 85.1 | 79.9 | 78.4 |
| | HcCNN [31] | 85.6 | 78.3 | 77.6 | 81.3 | 81.9 | 82.6 | 89.8 | **82.7** | 82.5 |
| | Ours | **87.6** | **87.5** | **81.2** | **83.6** | 79.0 | **83.1** | **90.8** | 75.4 | **83.5** |
| **Prec** | Inception-unbalanced [13] | 66.2 | **63.5** | **64.0** | 56.8 | 71.7 | 62.8 | 77.1 | 0.0 | 57.8 |
| | Inception-balanced [13] | 62.2 | 61.9 | 56.0 | 53.8 | 58.9 | 66.9 | 67.1 | 60.0 | 60.9 |
| | Inception-combined [13] | 65.3 | 61.6 | 52.7 | 57.8 | 56.5 | 70.5 | 63.0 | 30.8 | 57.3 |
| | EmbeddingNet [29] | **68.3** | 52.5 | 58.5 | 61.5 | 76.8 | 70.8 | 89.5 | 36.8 | 64.4 |
| | TripleNet [30] | 61.8 | 59.6 | 61.7 | 54.7 | 77.7 | 65.7 | 90.6 | 64.3 | 67.0 |
| | HcCNN [31] | 62.8 | 62.3 | 52.4 | 65.1 | **81.6** | 69.6 | **91.9** | 50.0 | 67.0 |
| | Ours | 65.6 | 48.4 | 50.4 | **82.3** | 73.5 | **74.9** | 67.4 | **100** | **70.3** |
| **Sens** | Inception-unbalanced [13] | 44.6 | 35.5 | 34.0 | 57.3 | 31.1 | 67.8 | 49.3 | 0.0 | 40.0 |
| | Inception-balanced [13] | 55.4 | 41.9 | 50.0 | **67.7** | 62.3 | 66.1 | 65.3 | 10.0 | 52.3 |
| | Inception-combined [13] | **61.4** | 48.4 | 51.1 | 59.7 | 66.0 | 62.1 | 77.3 | 13.3 | 54.9 |
| | EmbeddingNet [29] | 40.6 | 33.3 | 51.1 | 60.5 | 96.2 | 64.4 | **96.3** | 23.3 | 58.2 |
| | TripleNet [30] | 46.5 | 33.3 | 39.4 | 61.3 | **97.9** | 67.2 | 90.0 | **30.0** | 58.2 |
| | HcCNN [31] | 58.4 | 40.9 | 35.1 | 55.7 | 95.2 | **80.2** | 92.2 | 20.0 | **59.7** |
| | Ours | 59.0 | **77.5** | **67.0** | 39.2 | 21.9 | 70.1 | 69.9 | 3.6 | 51.0 |
| **Spec** | Inception-unbalanced [13] | 92.2 | **93.7** | **94.0** | 80.1 | 95.5 | 67.4 | 96.6 | **100** | **89.9** |
| | Inception-balanced [13] | 88.4 | 92.1 | 87.7 | 73.4 | 84.1 | 73.4 | 92.5 | 99.5 | 86.4 |
| | Inception-combined [13] | 88.8 | 90.7 | 85.7 | 80.1 | 81.3 | **78.9** | 89.4 | 97.5 | 86.6 |
| | EmbeddingNet [29] | **93.5** | 90.7 | 88.7 | 82.7 | 20.8 | 78.4 | 52.0 | 96.7 | 75.4 |
| | TripleNet [30] | 90.1 | 93.0 | 92.4 | 76.8 | 23.6 | 71.6 | 60.0 | 98.6 | 75.8 |
| | HcCNN [31] | 88.1 | 92.4 | 90.0 | 86.3 | 41.5 | 71.6 | 65.3 | 98.4 | 79.2 |
| | Ours | 89.5 | 79.0 | 80.3 | **95.8** | **96.8** | 78.8 | 91.0 | **100** | 88.9 |

| | | No CELM | +CELM | +CELM +GRM-DC |
|------|------|---------|-------|---------------|
| DIAG | BCC | 93.8 | **94.2** | 92.8 |
| | NEV | 82.8 | **87.7** | 86.8 |
| | MEL | **88.3** | 86.2 | 87.6 |
| | MISC | 88.3 | **89.2** | 88.8 |
| | SK | 71.4 | 73.3 | **79.8** |
| PN | TYP | 76.0 | 79.7 | **80.1** |
| | ATP | 83.7 | 85.4 | **87.5** |
| STR | REG | 78.1 | 81.2 | **84.9** |
| | IR | 80.2 | 77.1 | **81.2** |
| PIG | REG | 79.6 | 80.2 | **81.1** |
| | IR | 79.9 | 80.5 | **83.6** |
| RS | PRS | 74.0 | 75.4 | **79.0** |
| DaG | REG | 73.9 | 75.4 | **78.6** |
| | IR | 79.9 | 79.6 | **83.1** |
| BWV | PRS | 89.6 | 90.0 | **90.8** |
| VS | REG | 78.6 | **82.5** | 80.7 |
| | IR | 74.1 | **81.5** | 75.4 |
| Avg | | 80.7 | 82.3 | **83.6** |

Note: **Bold** indicates highest performance for each class.

| | Average AUC |
|------|-------------|
| +CELM+GRM-Separate | 82.8 |
| +CELM+GRM-Fused | 81.7 |
| +CELM+GRM-INV | 83.0 |
| +CELM+GRM-CD | 82.0 |
| +CELM+GRM-DC | **83.6** |

We examined the attention coefficients in GRM-DC to understand the strength of each relationship (Fig. S3 and Table S6). Notably, the average intermodality coefficients were relatively high. For both modalities in the first layer, coefficients of the edges directed from the DIAG node to the other categories were higher than corresponding self-attention coefficients. This disparity was less marked in the second layer for clinical and reversed for dermoscopy, where self-attention coefficients were substantially larger for all 7PC categories except DIAG.

The robustness of our GIIN to variations in the view and noise of test images is presented in Supplementary Table S7. The model was largely robust to the augmentations. Only the addition of random Gaussian noise to dermoscopy images caused a notable drop in test performance. The categories that were affected involved fine dermoscopic visual details: pigment network (PN), streaks (STR), and dots and globules (DaG).

### C. Application to Different Backbones

The results in Supplementary Table S8 indicate that our proposed modules consistently improved the classification results across different backbones by an average of 2.4 AUC. Performance on most of the classes were improved. Note that although the average AUC of ResNet-34 with our modules was higher than that of ResNet-50 with our modules, its overall sensitivity, specificity, and precision were inferior by a larger margin, hence we selected ResNet-50 as our main backbone network.

## IV. DISCUSSION

Our main findings are that: i) the ablation study showed that our CELM and GRM improved the performance of the baseline model and the GRM-DC configuration outperformed other topological variants; ii) our proposed GIIN had superior performance when compared to the state-of-the-art methods at classifying melanoma and the 7PC categories; and iii) the benefits introduced by our modules were consistent among various CNN backbones.

### A. Comparison to the State-of-the-Art

Our GIIN employed a novel strategy to combine multimodal information. Previous methods combined the modalities via the simple approaches of concatenation or averaging predicted probabilities [13, 29, 30]. HcCNN [31] used a multimodal branch to combine intermediate features at various scales from both modalities. None of the previous state-of-the-art methods leveraged the intercategorical relations within the 7PC. In contrast, the GIIN explicitly exploits the intrinsic relationships, and adaptively assigns weights such that the appropriate amount of importance can be placed on each modality per category. Our results show that this is superior to alternative approaches (Table 1). We note that while we do concatenate node features after the GRM, it occurs after intermodal processing, when complementary features have already been extracted and their relationships have been processed.

In the comparison results in Table 1, our GIIN showed some skewed behaviour for sensitivity, specificity, and precision. For example, its low sensitivity and high specificity for vascular structures (VS) indicate bias towards the negative class. We attribute this behaviour to the substantial class imbalance in the dataset (Table S1) [13]. There were relatively fewer examples of the positive classes, e.g., out of all 1,011 cases, there were only 71 examples of the positive (irregular, IR) class for VS, and 252 cases of melanoma. Class imbalance is a well-known issue in machine learning that can result in overfitting to the majority class. We attribute the low sensitivity of all methods to this issue. Kawahara et al. [13] attempted to alleviate such adverse effects by balancing training class examples with the Inception-balanced approach, though with mixed success. We did not use any class-balancing techniques in our experiments as our focus was the effectiveness of our modules. Despite this, our GIIN outperformed the state-of-the-art in terms of the mean of the average sensitivity, specificity, and precision.

### B. Ablation Study

In the ablation study we showed that both our modules improved the average classification performance across the attributes consistently for all three datasets (Table 2 and Table

S4). Though the main purpose of the CELM was to produce node embedding vectors for the categories, the module also improved the overall performance of the model due to the auxiliary learning procedure, which contributed to the individual losses for each modality, and thus provided further guidance to the intermediate features during learning. The CELM is essential for the GRM to learn successfully, as it conduces the general CNN features to specialise for each category so that the node representations of the GRM are meaningful. Unlike previous GCN-based multilabel approaches [36-38, 40, 41], our CELM does not use pretrained or predetermined label embeddings. Rather, it automatically learns with the rest of the model to transform general CNN features of the skin lesion into meaningful embeddings. Furthermore, the CELM does not require extra guidance other than the ground truth labels available from the dataset, such as semantic or disease knowledge.

The results for different topologies of the GRM (Table 3 and Table S3) demonstrated the power of our proposed structure at exploiting intercategorical and intermodal relations. The GRM-Separate variant examined the effectiveness of the star-shaped structure without blending features between modalities. The first layer of the module considers the 7PC categories for diagnosis, while each category also receives features from diagnosis. The second layer considers subtler relations between the 7PC categories through second-order representations. The results of GRM-Separate indicate that it is beneficial to exchange and share relevant features and relationships between categories according to the scheme of the 7PC. GRM-Fused attempted to integrate multimodal information via a concatenation of the vector embeddings of the modalities for each category. However, its poor performance suggests that it failed to leverage the complementary information of each modality, and that using a single modality graph structure is deleterious.

The purpose of the other three GRM variants—GRM-CD, GRM-INV, and GRM-DC—was to investigate the optimal directionality of intermodal edges to exploit complementary multimodal features. The direction of the edges between dermoscopy and clinical features determine the relative dominance of either modality (i.e., which modality has greater 'priority'), as the final outputs of the GRM contains 'unmixed' features from the more dominant modality and the 'intermixed' features from both modalities. As expected, prioritising clinical over dermoscopy in GRM-CD was slightly detrimental to performance (Table 3), since clinical images contain macroscopic information that are not sufficient for diagnosis [7-10]. This is also evidenced by single modality results in Table S5, which show weaker performance with using only clinical images compared to dermoscopy. By using intermodal edges directed from clinical to dermoscopy images, the dermoscopy features were effectively corrupted with less accurate information, thereby reducing performance. This topology is therefore not suitable to the data, though its relatively high performance on irregular (IR) pigmentation (PIG) (Table S3) may suggest applicability to features that can be perceived at a larger scale. The performance of GRM-INV suggests that it is approximately an average of the effects of GRM-CD and GRM-DC. The inverse edges exchange features between the modalities and result in both the detrimental and beneficial effects of prioritising the weaker and stronger modalities, respectively.

Our proposed configuration, GRM-DC, was able to leverage the benefits of intercategorical relations and complementary multimodal features to improve classification, outperforming all GRM alternatives (Table 3). While the 7PC does not specify relationships between clinical and dermoscopy image features, our module is able to leverage useful relationships to improve performance. By using intermodal edges directed from dermoscopy to clinical, GRM-DC processes isolated dermoscopy features, and uses dermoscopy to supplement the broad clinical features, thereby using both modalities whilst recognising that the more advanced imaging technique of dermoscopy is the more valuable modality. Categories that are most clearly visible in dermoscopy benefitted from this topology, such as atypical (ATP) pigment network (PN) (Table 2). Moreover, we demonstrate that this configuration outperforms single modality inputs with or without CELM and GRM in Table S5 as further evidence that our proposed module uses complementary information from both modalities.

The attention coefficients of the GATs in our GRM (Fig. S3 and Table S6) revealed that the contributions of dermoscopy towards clinical features were relatively large, which further reinforces the importance of the DC edges. The coefficients also imply that it is more difficult to discern each 7PC category compared to directly classifying diagnosis, as the contributions from diagnosis towards the other categories were relatively larger than their self-contributions. This suggestion is consistent with previous reports of poorer classification performance on individual skin lesion attributes (such as streaks and pigmented network) than diagnosis [42]. By leveraging intercategorical relationships, our modules boosted the performance of all categories when compared to the baseline.

Our GIIN is robust to variations in the view of the input images (Table S7), which is beneficial in clinical settings where the field-of-view of clinical images are not standardised. This robustness can be attributed to the use of data augmentation during training, and to the convolutional and GAP layers that provide invariance to such changes. Regarding the addition of Gaussian noise, this augmentation technique is likely to be more detrimental to fine details such as pigment network (PN), streaks (STR), and dots and globules (DaG). As these categories are more prominent in dermoscopy, performance was reduced when Gaussian noise was added to the dermoscopic images.

## C. Application to Different Backbones

We showed that our CELM and GRM modules can be applied to different backbone CNNs. This relates to our designing the CELM so that it could operate on a general set of CNN features from the input images and transform them into category-specific representations that are suitable for the GRM. The consistent enhancement of classification performance across the different categories for various backbones that was due to our modules (Table S8) demonstrates their robustness and the value of exploiting intercategorical and intermodal relationships.

### D. Future Work

More layers can be added into the GRM to investigate whether it is beneficial to include relationships that are higher than second order. Alternative topologies that emphasise additional associations between categories may be assessed depending on reported evidence.

Threshold values need to be chosen for the predicted probabilities if the model was to be deployed. The risks for false positive and false negative classifications need to be balanced. For example, in a general practice setting, a low threshold for positive classes would be feasible for an early screening tool to recommend follow-up visits to specialists to filter out false positives.

## V. CONCLUSION

We introduced a graph-based relational module to exploit intercategorical relationships and complementary multimodal features, as they had not been fully leveraged by previous studies. Our results showed that classification performance can be improved by using the inherent relations within the 7PC. Performance is enhanced the most when detailed dermoscopic features and broader macroscopic information from clinical images are leveraged by intermodal edges directed from the dermoscopy to clinical modalities, compared to other directions of intermodal edges. Our category embedding learning module can automatically produce vector embeddings for each category from general CNN features of the lesion to assist intercategorical and intermodal learning. Our GRM and CELM modules consistently improved the classification performance of various baseline models. Our proposed GIIN outperformed state-of-the-art methods on a public skin lesion dataset.